\documentclass[runningheads]{llncs}
\usepackage{amsmath,graphicx}

\usepackage{amssymb}
\usepackage{adjustbox}
\usepackage{hyperref}
\usepackage{textcomp,subfigure,multirow,upgreek,color,microtype}
\newcommand{\mname}[0]{{\bf AnonyGAN}}

\begin{document}\sloppy
\title{Graph-based Generative Face Anonymisation with Pose Preservation}
%
%

\author{Nicola Dall'Asen\inst{1,2} 
\and Yiming Wang\inst{3}\orcidID{0000-0002-5932-4371} 
\and Hao Tang\inst{4}
\and Luca Zanella\inst{3}
\and Elisa Ricci\inst{2,3}\orcidID{0000-0002-0228-1147}}

%
\authorrunning{N. Dall'Asen et al.}
%

\institute{
    University of Pisa, Pisa, Italy \\
    \email{nicola.dallasen@phd.unipi.it}
    \and University of Trento, Trento, Italy
    \and Fondazione Bruno Kessler, Trento, Italy 
    \and {ETH Zürich, Zürich, Switzerland}\\
}
\maketitle              
\begin{abstract}
We propose AnonyGAN, a GAN-based solution for face anonymisation which replaces the visual information corresponding to a source identity with a condition identity provided as any single image. With the goal to maintain the geometric attributes of the source face, i.e., the facial pose and expression, and to promote more natural face generation, we propose to exploit a Bipartite Graph to explicitly model the relations between the facial landmarks of the source identity and the ones of the condition identity through a deep model. We further propose a landmark attention model to relax the manual selection of facial landmarks, allowing the network to weight the landmarks for the best visual naturalness and pose preservation. Finally, to facilitate the appearance learning, we propose a hybrid training strategy to address the challenge caused by the lack of direct pixel-level supervision. We evaluate our method and its variants on two public datasets, CelebA and LFW, in terms of visual naturalness, facial pose preservation and of its impacts on face detection and re-identification. We prove that AnonyGAN significantly outperforms the state-of-the-art methods in terms of visual naturalness, face detection and pose preservation. Code and pretrained model are available at \url{https://github.com/Fodark/anonygan}. 
\end{abstract}

%

%
%
%
\section{Introduction}
\label{sec:intro}

In the era of deep learning, the availability of large scale data has undoubtedly brought technological advances. However, the very same fact has also fostered the growing concern regarding privacy issues. 
Visual privacy preservation is mostly achieved via video redaction methods by obfuscating the personally identifiable information (PII) of a data subject, whose face is often the most identity-informative part. Classic face anonymisation techniques, e.g., blurring \cite{du2019efficient} or pixelation \cite{gerstner2013pixelated}, can effectively remove PII. However, this comes at a high cost of degrading other vision-related tasks, in particular for the action/emotion recognition where the poses play an essential role. 

Thanks to recent advances in generative adversarial networks (GANs) \cite{goodfellow2014generative}, several anonymisation solutions have been proposed to generate {\em natural-looking} faces that correspond to different identities \cite{sun2018natural,face_deidentification_iccv2019,SimSwap_2020,li2019faceshifter}, while preserving the original facial poses using the landmarks as the guidance \cite{sun2018hybrid,ciagan_cvpr2020,deepprivacy2019}.
%
Yet, it is challenging to produce realistic images in this context due to the lack of ground-truth images in real-world, i.e., images of different persons with the same facial pose. This also draws a fundamental distinction to the pose-guided image generation task \cite{grigorev2019coordinate,esser2018variational,tang2020bipartite}, whose ground-truth images, i.e. images of the same person with varying poses, are available to provide direct pixel-level supervision. While for the pose preservation, the main challenge lies in the relation reasoning between the condition pose and the source pose, where graph-based modelling has demonstrated its strengths in such geometric reasoning \cite{tang2020bipartite}. Finally, the landmarks choice impacts the visual naturalness, pose preservation and anonymisation, which is often heuristically handled with no optimality guaranteed \cite{ciagan_cvpr2020,deepprivacy2019}.

\begin{figure}[!t] \small
	\centering
	\includegraphics[width=.9\linewidth]{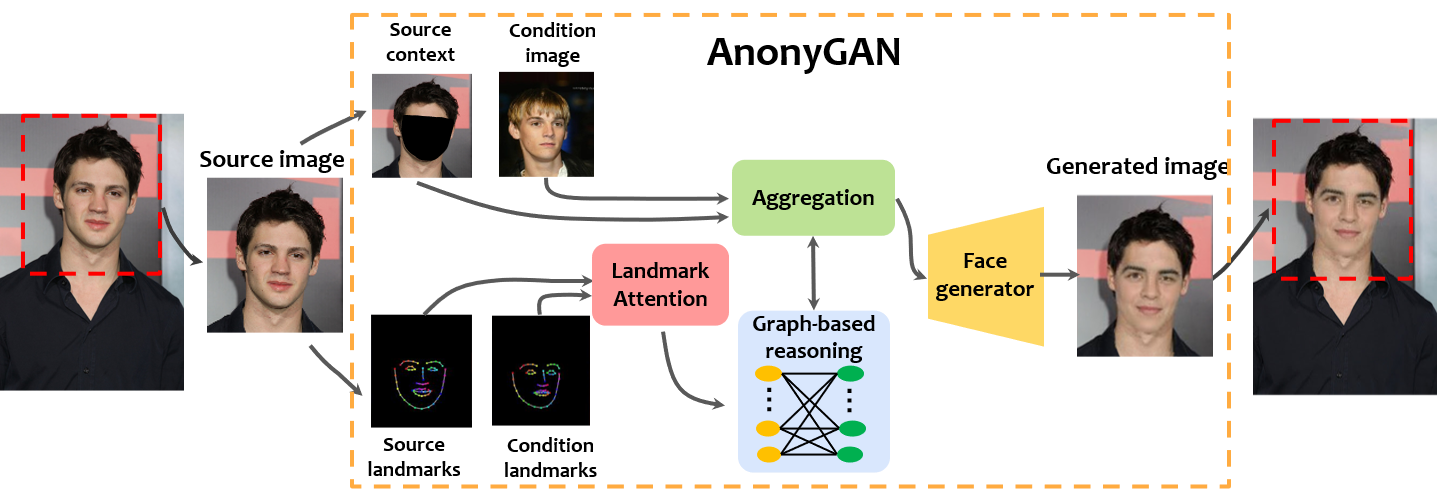}
	\vspace{-0.4cm}
	\caption{\mname~anonymises faces by generating similar faces to {\em any} condition image while preserving the facial pose of the source image, with a novel {\em landmark attention} model and a face generator with {\em graph-based landmark-landmark reasoning}.} 
	\label{fig:teaser}
\end{figure}

To address the above-mentioned challenges, we propose a novel graph-based GAN architecture, \mname, to perform landmark-guided face anonymisation (see Figure \ref{fig:teaser}). Our network takes as input a \emph{context} image and a condition image, as well as the facial landmarks extracted from the source and target images. The {\em context} image is the source image with the face (excluding the forehead) masked out, and it provides the necessary contextual information, e.g. the skin tone and the background, for naturally blending the generated faces to the source image. 
In order to improve the pose preservation, we propose to first disentangle the geometrical reasoning from the appearance, where the source landmarks and the condition landmarks are modelled as a bipartite graph with Graph Convolution Networks (GCNs). The appearance of the condition image is then aggregated to the pose reasoning module to generate more natural faces with the source facial pose preserved. 
Moreover, we introduce a novel landmark attention model to allow the network to automatically learn the importance of the facial landmarks, avoiding sub-optimal manual decisions.
Finally, we propose a novel hybrid training strategy to address the training difficulty caused by the lack of ground-truth images. We form the source and condition pairs using both the same image and different images to facilitate the appearance learning. The former provides direct pixel-level supervision, while the latter applies a weak context-level supervision by exploiting the high-level features extracted from the appearance discriminator.
We validate \mname~on two public datasets, i.e., CelebA \cite{celeba_iccv2015} and LFW \cite{lfw_2008}, and demonstrate that \mname~can greatly improve the generated faces in terms of visual naturalness and pose preservation compared to baselines, i.e., blurring and pixelation, and the state-of-the-art method \cite{ciagan_cvpr2020}. 
To summarise, our main contributions are listed below:
\begin{itemize}
	\item We propose a novel GAN-based architecture, \mname,~to perform landmark-guided face anonymisation, achieving the {\em highest visual quality} with the {\em best pose preservation} on two benchmark dataset. 
	\item We propose to exploit a graph formulation on the landmarks of the source and condition face images to perform geometric reasoning using GCNs, and prove its effectiveness in improving the pose preservation.
	\item We propose a landmark attention model to automatically weight the facial landmarks, achieving a higher visual quality and perception performance.
	\item We propose a hybrid training strategy with a strong pixel-level and a weak context-level supervision to address the training without ground-truth images, achieving the best visual quality.
\end{itemize}
\section{Related Work} \label{sec:sota}
We discuss recent face obfuscation techniques for anonymising visual data and briefly cover related works addressing pose-guided image generation.

\noindent \textbf{Visual Anonymisation} often refers to irreversible obfuscation techniques for removing PII of the data subject in visual content, a.k.a. de-identification in some works \cite{face_deidentification_iccv2019}. Many works anonymise visual data by obfuscating the faces, the most privacy-concerning content, using classic techniques, such as blurring via filters \cite{du2019efficient} or pixelation by enlarging the pixels \cite{gerstner2013pixelated}. 
Such methods remove the most identifiable visual information, but greatly compromise other perception tasks, such as object detection and action recognition. 
Recently, with the progresses in GANs, face anonymisation techniques have advanced by generating realistic faces of a different identity, leaving intact most of the non-identifiable visual and geometrical attributes \cite{deepprivacy2019,face_deidentification_iccv2019,sun2018natural,sun2018hybrid,ciagan_cvpr2020}. Sun {\em et al.} \cite{sun2018natural} proposed a two-step face inpainting technique for anonymisation by first generating the 68 facial landmarks, and then synthesising the faces guided by the landmarks. With a blurred face as condition, the generated faces have a rather high visual quality, yet resemble to the original face. DeepPrivacy \cite{deepprivacy2019} exploits the generator of StyleGAN \cite{stylegan_cvpr2019} to generate a face of a fake identity, conditioned both on the context image with the face masked out and on 7 facial landmarks. The generated faces are limited in anonymisation and pose preservation. Conditional GANs are also proposed to explicitly control the identity on the generated faces \cite{sun2018hybrid,ciagan_cvpr2020}. Recently, CIAGAN \cite{ciagan_cvpr2020} has introduced an identity discriminator to enforce the generated faces to be different from the source image, achieving a better anonymisation performance, while the visual naturalness and pose preservation are not yet satisfactory although the subset of landmarks are carefully chosen. Moreover, CIAGAN cannot be easily applied to unknown condition identity, as the condition identities are encoded within the network during the training. 

\mname~takes any condition image as a reference for appearance, and improves the visual naturalness and facial pose preservation by first reasoning on landmark to landmark relations, and then generating the image aggregating the appearance features. Moreover, the choice of facial landmarks impacts both the image quality and the perception task with a certain trade-off. This aspect has not been properly addressed in the literature yet \cite{ciagan_cvpr2020}. Instead, in this work we propose a landmark attention model to allow the network to learn the relative importance among the landmarks for face anonymisation. 

\noindent \textbf{Pose-Guided Image Generation} is a related task to visual anonymisation where the main challenge is due to the pose deformation between the source and the target image. Modelling the relations between the source pose and the target pose is the key to solve this challenging task.
Existing methods such as \cite{balakrishnan2018synthesizing,siarohin2018deformable,tang2019cycle,albahar2019guided,zhu2019progressive,chan2019everybody,liang2019pcgan,liu2019liquid} are based on the stacking of several convolutional layers, which can only leverage the relations between the source pose and the target pose locally. For instance, Zhu \emph{et al.}~\cite{zhu2019progressive} proposed a Pose-Attentional Transfer Block, in which the source and target poses are simply concatenated and then fed into an encoder to capture their dependencies. Different from existing methods for modelling the relations between the source and target poses in a localised manner, BiGraphGAN \cite{tang2020bipartite} reasons and models the crossing long-range relations between different features of the source pose and target pose in a bipartite graph, which are then used for the person image generation process. In general, for pose-guided image generation, the ground-truth images of the same person with various poses are available. 
In contrast, there is no set of images of different identities with the exact same pose, thus making the face anonymisation a more challenging problem due to the lack of direct pixel-level supervision.  

\section{Proposed Method}
\label{sec:method}
\begin{figure}[t!] \small
	\centering
	\includegraphics[width=.9\linewidth]{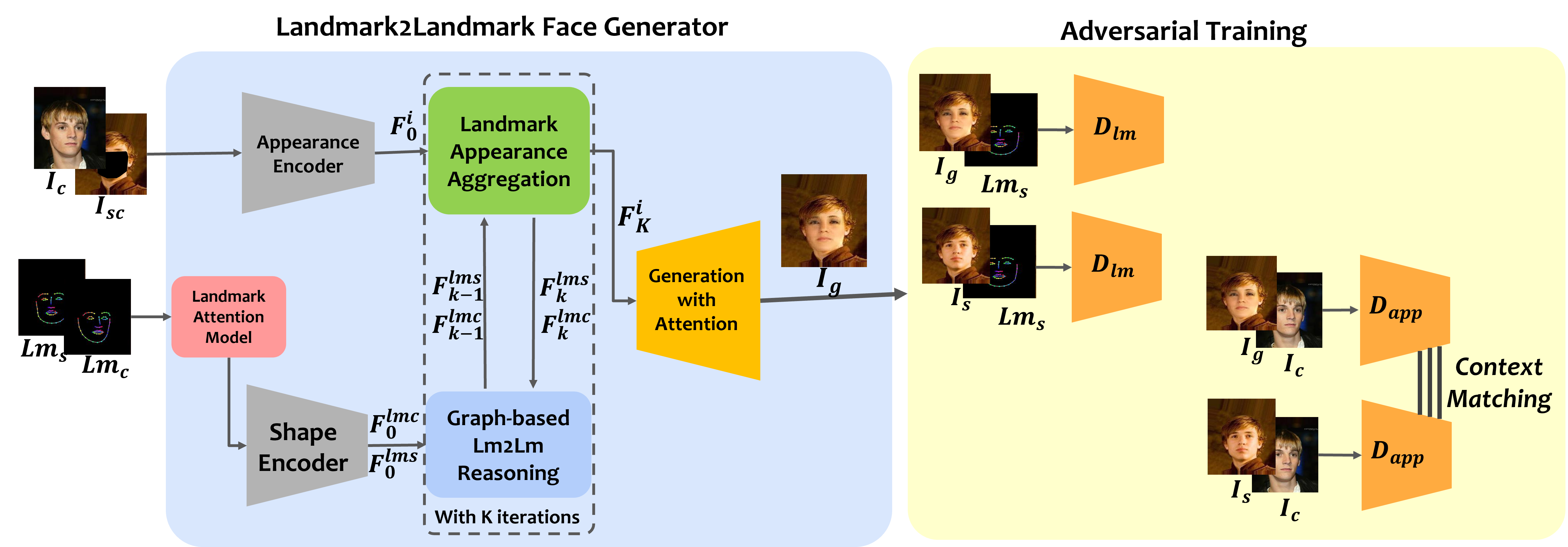}
	\vspace{-0.4cm}
	\caption{Network architecture of \mname. The condition image ${I}_{c}$ and source context image $I_{sc}$ are encoded with an appearance encoder, while the source and condition landmarks ${Lm}_s$ and ${Lm}_c$ are encoded by a shape encoder. The shape codes ${F}_{0}^{lmc}$ and ${F}_0^{lms}$ are passed to the graph-based landmark reasoning model, and iteratively aggregated with the appearance code for K iterations. The final appearance code ${F}_{K}^{i}$ is used for the face generation. The adversarial training is performed with two discriminators: The Appearance Discriminator ${D}_{app}$ optimises for naturally blending the facial attributes of the target into the source image, and the Landmark Discriminator ${D}_{lm}$ optimises for preserving the source pose.  
	}
	\label{fig:method}
\end{figure}

The proposed \mname, as illustrated in Figure \ref{fig:method}, aims to anonymise a source image of any identity $I_s$ by replacing the face with any condition identity provided as a condition image $I_c$ while preserving the facial pose of source image $Lm_s$. Both the source image $I_s$ and the condition image $I_c$ are preprocessed to extract their facial landmarks $Lm_s$ and $Lm_c$, respectively, for the graph-based landmark-landmark reasoning. Both facial poses are defined by 68 landmarks and encoded as 68-channel images as in \cite{SAGONAS20163} with a channel per landmark. In order to generate the face with a consistent context of the source image, in particular, for the skin tone, we also input to the network the context of the source image $I_{sc}$, i.e., the background image with the face masked out by the contour defined by the facial landmarks.


The main components of our network are: a {\em Landmark Attention Model} that takes as input $Lm_s$ and $Lm_c$ defined by the full 68 landmarks, and learns to weight the landmarks to optimally trade-off between the face naturalness and re-identification performance; a {\em Landmark2Landmark Face Generator} that follows a similar architecture of the generator of BiGraphGAN \cite{tang2020bipartite}. We use GCNs to learn the spatial relations between $Lm_s$ and $Lm_c$, aggregated with the visual features extracted from the condition image $I_c$. With several iterations of the landmark appearance aggregation, the final image $I_g$ is generated with an attention model. 
For {\em adversarial training}, we adopt a {\em Landmark Discriminator}  $\mathbf{D}_{lm}$ and an {\em Appearance Discriminator} $\mathbf{D}_{app}$.  $\mathbf{D}_{lm}$ is designed to preserve the facial pose of the source image, while $\mathbf{D}_{app}$ is designed to preserve the appearance of the condition face with the skin tone matched to the source face.
We will provide more details on the {\em Landmark Attention Model}, {\em Landmark2Landmark Face Generator} and {\em Adversarial Training} in the following sections.

\subsection{Landmark Attention Model}
\label{sec:method:lam}

The landmark attention model is designed to learn the optimal weighting strategy on the landmarks to achieve jointly the best visual naturalness and pose preservation. We first concatenate the 68-channel landmark maps of both the source facial pose and the condition facial pose, and feed it to an Efficient Channel Attention module \cite{wang2020eca} formulated as:
\begin{equation}
\begin{aligned}
    \omega = \sigma({{\it Conv1D}}_{J} (GAP(Concat(Lm_{s}, Lm_{c})))),
\end{aligned}
\end{equation}
where $\sigma(\cdot)$ is the Sigmoid function, ${\it Conv1D}_{J}(\cdot) $ is 1-D convolution with kernel size $J$, $GAP(\cdot)$ represents the operation of the channel-wise Global Average Pooling, and $Concat(\cdot)$ is the concatenation operation.


\subsection{Landmark2Landmark Face Generator}
\label{sec:method:generator}

The landmark2landmark face generator follows the architecture of \cite{tang2020bipartite} that iteratively reasons the relations between the source landmarks $Lm_s$ and the condition landmarks $Lm_c$ following a bipartite graph formulation, and aggregates with the appearance feature of the condition image $I_c$ and the context of the source image $I_{sc}$. The final aggregated feature is used for the landmark-guided face generation with the condition identity.

The condition image $I_c$ and the source context image $I_{sc}$ are first concatenated and fed to an appearance encoder to generate the initial appearance code $F_{0}^{i}$, while the $Lm_s$ and $Lm_c$ after the Landmark Attention Model are passed to a shape encoder to obtain the shape codes $F_{0}^{lms}$ and $F_{0}^{lmc}$, respectively. The shape codes are then fed to a graph-based landmark-to-landmark reasoning model in a bipartite graph via GCNs to update the shape codes, which are then fed to the Landmark Appearance Aggregation model to synchronise the updates in both appearance and shape codes. Such operation of landmark-to-landmark reasoning and appearance aggregation is performed {\em iteratively} to form a thorough reasoning from low to high level. 
At the $K$-th iteration of the graph-based landmark-landmark reasoning and appearance aggregation, the final appearance code $F_K^{i}$ is passed to both an image decoder to generate the intermediate result $\tilde{I}_g$, and an attention decoder to produce the attention mask $A_i$, a one-channel attention mask with the pixel value between 0 to 1. The final generated image is obtained by $I_g = I_c \otimes  A_i + \tilde{I}_g \otimes (1 - A_i)$,
where $\otimes$ denotes element-wise product.


\subsection{Adversarial Training}
\label{sec:method:trainig}
Two discriminators are designed for the adversarial training. Specifically, 
the {\em Landmark Discriminator} $\mathbf{D}_{lm}$ is fed with the landmark-image pairs of the source $\{Lm_s, I_s\}$ and the generated $\{Lm_s, I_g\}$ to encourage the generation of similar facial pose of the source image. 
The {\em Appearance Discriminator} $\mathbf{D}_{app}$ takes the source-condition image pair $\{I_s, I_c\}$ and the generated-condition pair $\{I_g, I_c\}$ to guide the face generation with the facial attributes of the condition face within the same context of the source image by coupling $I_s$ and $I_g$ with $I_c$.
In order to facilitate the appearance learning, we train the Appearance Discriminator in a hybrid manner with the source-condition image pairs composed by either the same image or a couple of images with different poses and contexts. When the source and condition images are the same, i.e. $I_s = I_c$, we apply a L1 loss on $I_g$ and $I_s$ to provide the strong pixel-level supervision. When the source and condition images are different, we apply a weak supervision for {\em Context Matching} by enforcing similar high-level features of $I_g$ and $I_s$ that correspond to skin tone and background.


We employ several losses to drive the network learning. 
We train $D_{app}$ with the adversarial loss $L_{app}$:
\begin{equation}
    L_{app} = \min\limits_{G}\max\limits_{D_{app}} \mathbb{E}[log(\mathbf{D}_{app}(I_{s}, I_{c}))] + \mathbb{E}[1 - log(\mathbf{D}_{app}(I_{g}, I_{c}))].
\end{equation}

The landmark discriminator $D_{lm}$ is trained with $L_{lm}$, driving the generator towards the correct pose:
\begin{equation}
\resizebox{.8\linewidth}{!} {
$L_{lm} = \min\limits_{G}\max\limits_{D_{lm}} \mathbb{E}[log(\mathbf{D}_{lm}(I_{s}, Lm_{s}))] + \mathbb{E}[1 - log(\mathbf{D}_{lm}(I_{g}, Lm_{s}))].
$}
\end{equation}

Moreover, when the source and condition pair is of different images, the {\em Context Matching} supervision is realised through the weak Feature Matching (wFM) loss \cite{SimSwap_2020} defined as:
\begin{equation}
L_{wFM}(\mathbf{D}_{app}) = \sum_{i=m}^{M} \frac{1}{N_{i}} ||\mathbf{D}^{(i)}_{app}(I_{g}, I_{c}) - \mathbf{D}^{(i)}_{app}(I_{s}, I_{c})||_{1},
\end{equation}
where $D^{(i)}_{app}$ denotes the feature map produced by the \textit{i}-th layer of the discriminator $D_{app}$; $N_{i}$ is the number of elements in the feature map produced by the \textit{i}-th layer; \textit{m} is the first layer from which the weak feature matching loss computation starts; and \textit{M} is the total number of layers of the discriminator $D_{app}$. 
When the condition and source pair is composed of the same image, we provide the generator with the direct pixel-level supervision using the image reconstruction loss $L_{Recon} = ||I_{g} - I_{s}||_1$.

The {\em final loss} function is expressed as a weighted sum of the above-mentioned loss terms, with the weighting parameters empirically set. 
\section{Experiments}
\label{sec:exp}
We evaluate our proposed method \mname~and its variants in comparison with state-of-the-art methods using two public face datasets. We validate the effectiveness of our design choices by evaluating the generated faces in terms of the visual naturalness, pose preservation, face detection and anonymisation. 

\noindent\textbf{Datasets.} 
Our model is evaluated on two benchmark datasets created for face-related computer vision tasks, i.e., CelebA \cite{celeba_iccv2015} and Labeled Faces in the Wild (LFW) \cite{lfw_2008,LFWTechUpdate}. 
\textbf{CelebA} is a large-scale dataset of 202,599 images with different poses and backgrounds of 10,177 celebrities. 
\textbf{LFW} is composed of 13,233 face images collected from the web covering 5,749 identities with 1,680 people having two or more images. 
We use CelebA for both training and testing, while LFW is used for testing only. 
We follow the same train/test split of CelebA as in \cite{ciagan_cvpr2020}, where in total 1,563 identities with more than 30 images per identity are selected. 
The training set is composed of 1,200 identities, where 24,000 pairs are formed for the training. We use the images of the remaining 363 identities for testing, where each source image is paired with a condition image that is randomly selected from the images of the next identity as in \cite{ciagan_cvpr2020}. 
For the test set of LFW, we follow the protocol defined in \cite{face_deidentification_iccv2019}, where we form 6,000 pairs of images organised in 10 different folds with each folder containing half of the pairs corresponding to the same person, and the other half corresponding to different persons. 
We use Dlib \cite{dlib09} to preprocess the 68 facial landmarks, which are then used to define the mask area of each face.


\noindent\textbf{Evaluation Metrics.}
We evaluate the performance of our model in terms of visual quality, pose preservation, face detection, and face re-identification. The \textbf{visual quality} of generated faces is measured by the Fréchet inception distance (FID) \cite{heusel2017gans}, which 
calculates the distance between real and synthetic images in a feature space given by a specific layer of Inception Net. The lower the FID score is, the higher the quality of generated images\footnote{FID Implementation is taken from: \url{https://github.com/mseitzer/pytorch-fid}}. 
The \textbf{pose preservation} is measured by the $L1$ distance between the detected 68 landmarks and the ground truth landmarks, normalised by the inter-ocular distance \cite{pose_metric}. 
The model should generate faces that minimally impact the visual detection task. We evaluate the \textbf{face detection} performance on the generated images using two face detection algorithms, i.e., Dlib \cite{king2009dlib} and FaceNet \cite{schroff2015facenet}. A higher detection rate is more desired. Meanwhile, the generated faces should maximally prohibit the \textbf{face re-identification} for the best anonymisation. We report the rate of the correct matches of the generated faces and source faces using FaceNet without fine-tuning. In particular, for the test of LFW, we compute for each fold the re-identification rate, and report the mean and the standard deviation among all folds. A lower re-identification rate indicates a better face anonymisation.

\noindent\textbf{Implementation Details.}
We set the learning rate for the generator during training to $2*10^{-4}$ and the learning rate for discriminators to $2*10^{-6}$. 
%
We perform hybrid training using the condition and source image pairs with the same images and different images to facilitate the appearance learning. 
During training, we we set 75\% of pairs with $I_{c}=I_{s}$ for each batch. 

\noindent\textbf{Method Comparisons.} We compare \mname~ against a set of baselines and state-of-the-art GAN-based methods that are closely related to the face obfuscation task: 1) {\bf Blurring} applies a $101 \times 101$ Gaussian kernel to blur the face region. 2) {\bf Pixelation} uses a $10 \times 10$ pixelation mask on the face region. 
3) {\bf CIAGAN} \cite{ciagan_cvpr2020} is a state-of-the-art face anonymisation method based on conditional GAN, which accepts as input the context image and 29 facial landmarks of the source image and a conditioning ID, and generates a face matching both the context and the conditioning ID. We use the inference model provided by the authors for the evaluation. 

Moreover, in order to demonstrate the validity of our design choices and, in particular, the importance of our Landmark Attention (LA) and Context Matching (CM), we ablate several variants of \mname: 
1) \mname-$\mathbf{{(CM,LA)}^{-}}$ (68 lm) is trained without Landmark Attention and Context Matching, with the full 68 facial landmarks. 
2) \mname-$\mathbf{{(CM,LA)}^{-}}$ (29 lm) is trained without Landmark Attention (LA) and Context Matching (CM), but with 29 facial landmarks that are manually selected as in CIAGAN \cite{ciagan_cvpr2020}, in order to show the impact of the landmark choice.
3) \mname-$\mathbf{{(CM)}^{-}}$ (68 lm) is trained with Landmark Attention but without the Context Matching, with 68 facial landmarks, to prove the capability of the Landmark Attention module on automatically weighting the landmarks.
Finally, \mname (68 lm) is trained with both Landmark Attention and Context Matching, with all 68 landmarks, to justify the capability of Context Matching for improving visual fidelity.

 
\noindent\textbf{Result Discussion.} Table \ref{tab:comparison_celeba} presents the results of all methods evaluated on the test set of CelebA. Classic face obfuscation techniques, i.e. blurring and pixelation, reduce greatly the visual quality, thus hampering face detection. On the other hand, they achieve the lowest re-identification rate, thus the best anonymisation performance. The pose preservation metric is not available as the landmarks are not detectable from the anonymised faces.
The visual quality of the images generated by CIAGAN \cite{ciagan_cvpr2020} is in general inferior to our \mname~models as indicated by the FID metric. The unnaturalness of the generated faces leads to a lower detection rate with the standard face detector Dlib, but also confuses the face re-identification module, leading to a better anonymisation performance.
Moreover, our \mname~models significantly improve the pose preservation performance thanks to the graph-based geometric reasoning.

Interestingly, by choosing a subset of the landmarks (29 out of 68 landmarks) carefully as in CIAGAN, we observe that \mname-$\mathbf{{(CM,LA)}^{-}}$ (29 lm) improves the visual quality without impacting the pose preservation, compared to \mname-$\mathbf{{(CM,LA)}^{-}}$ (68 lm). 
Moreover, the result of \mname-$\mathbf{{CM}^{-}}$ (68 lm) in comparison to \mname-$\mathbf{{(CM,LA)}^{-}}$ (29 lm) shows that the introduction of Landmark Attention enables the network to learn the importance of landmarks automatically, achieving a better visual quality without impacting the pose preservation. Finally, \mname~with both Landmark Attention and Context Matching achieves the best results in terms of face visual quality, face detection and pose preservation, however with a slight compromise in the anonymisation performance compared to the variants and CIAGAN. These above-mentioned observations are also in line with the results evaluated on the LFW dataset as reported in Table~\ref{tab:comparison_lfw}.

\begin{table}[t] \small
	\centering
	\caption{Performance of proposed \mname~and its variants evaluated on the test set of CelebA, compared to baselines and state-of-the-art GAN-based method.}
	\begin{adjustbox}{max width=.9\linewidth}
    \begin{tabular}{lccccccc}
    \hline
    \multirow{2}{*}{} & \multirow{2}{*}{$FID \downarrow$} & \multicolumn{2}{c}{Face detection $\uparrow$} & \multicolumn{2}{c}{Face re-identification $\downarrow$} & \multirow{2}{*}{Pose$\downarrow$}\\
    & & dlib & Facenet &  Casia & VGG \\
    \hline
    Blurring & 95.13 & 4\% & 4\% & {\bf 0.07}\% & {\bf 0.02}\% & -\\
    Pixelation & 59.82 & 1\% & 28\% & 0.28\% & 0.12\%  & -\\
    \hline
    CIAGAN~\cite{ciagan_cvpr2020} (recomp.) & 37.94 & 96\% & \textbf{100}\% & 1.61\% & 0.51\%  & 1.44\\
    \hline
    \mname-$\mathbf{{(CM,LA)}^{-}}$ (68 lm) & 43.99 & \textbf{100}\% & \textbf{100}\% & 2.63\% & 0.58\%  & \textbf{0.16}\\
    \mname-$\mathbf{{(CM,LA)}^{-}}$ (29 lm) & 30.24 & \textbf{100}\% & \textbf{100}\% & 2.84\% & 0.66\%  & \textbf{0.16}\\
    \mname-$\mathbf{{CM}^{-}}$ (68 lm) & 26.12 & \textbf{100}\% & \textbf{100}\% & 2.70\% & 0.91\%  & \textbf{0.16}\\
    \hline
    \mname (68 lm) & \textbf{22.53} & \textbf{100\%} & \textbf{100\%} & 3.52\% & 1.60\%  & \textbf{0.16}\\
    \hline
\end{tabular}
\end{adjustbox}
\label{tab:comparison_celeba}
\end{table}

\begin{table}[!ht] \small
	\centering
	\caption{Performance of the proposed \mname~and its variants evaluated on the test set of LFW in comparison to the baselines and state-of-the-art GAN-based method.}
	\begin{adjustbox}{max width=.9\linewidth}
    \begin{tabular}{lccccccc}
    \hline
    \multirow{2}{*}{} & \multirow{2}{*}{$FID \downarrow$} & \multicolumn{2}{c}{Face detection $\uparrow$} & \multicolumn{2}{c}{Face re-identification $\downarrow$} & \multirow{2}{*}{Pose$\downarrow$}\\
    & & dlib & Facenet &  Casia & VGG \\
    \hline
    Blurring & 85.03  & 7\% & 17\%  & \textbf{0.03\% $\pm$ 0.07} & \textbf{0.02\% $\pm$ 0.05} & -\\
    Pixelation & 47.97 & 6\% & 22\% & \textbf{0.03\% $\pm$ 0.07} & \textbf{0.02\% $\pm$ 0.05} & -\\
    CIAGAN~\cite{ciagan_cvpr2020} (recomp.) & 18.37 & 99\% & \textbf{100}\% & 1.28\% $\pm$ 0.32 & 0.08\% $\pm$ 0.12  & 0.45\\
    \hline
    \mname-$\mathbf{{(CM,LA)}^{-}}$ (68 lm) & 44.41 & 99\% & \textbf{100}\% & 3.93\% $\pm$ 0.67 & 0.38\% $\pm$ 0.19  & 0.11\\
    \mname-$\mathbf{{(CM,LA)}^{-}}$ (29 lm) & 19.42 & 97\% & \textbf{100}\% & 4.18\% $\pm$ 0.45 & 0.35\% $\pm$ 0.31  & 0.11\\
    \mname-$\mathbf{{(CM)}^{-}}$ (68 lm) & 14.66 & \textbf{100}\% & \textbf{100}\% & 3.35\% $\pm$ 0.78 & 0.32\% $\pm$ 0.27  & 0.06\\
    \hline
    \mname (68 lm) & \textbf{8.93} & \textbf{100}\% & \textbf{100}\% & 3.55\% $\pm$ 0.94 & 0.48\% $\pm$ 0.29  & \textbf{0.05}\\
    \hline
\end{tabular}
\end{adjustbox}
\label{tab:comparison_lfw}
	\vspace{-0.3cm}
\end{table}

\begin{figure*}[!ht]
  \centering
  \begin{adjustbox}{max width=0.45\textwidth}
  
\begin{tabular}{c|ccccccc}
Condition & Source & CIAGAN\cite{ciagan_cvpr2020} & \mname (68 lm)\\ 
\hline
\includegraphics[width=2.5cm]{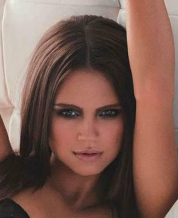} & \includegraphics[width=2.5cm]{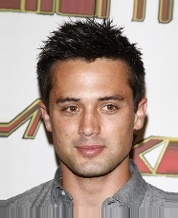} & \includegraphics[width=2.5cm, height=3.05cm]{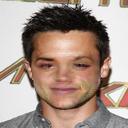} & 
\includegraphics[width=2.5cm]{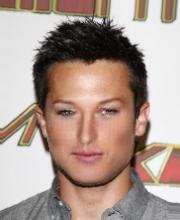}\\
\includegraphics[width=2.5cm]{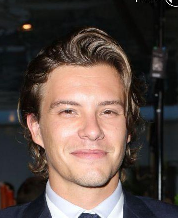} & \includegraphics[width=2.5cm]{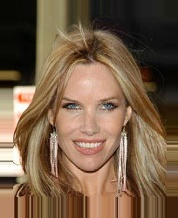} & \includegraphics[width=2.5cm, height=3.05cm]{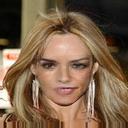} & 
\includegraphics[width=2.5cm]{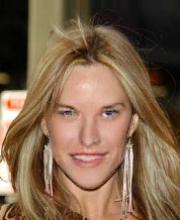}\\
\includegraphics[width=2.5cm]{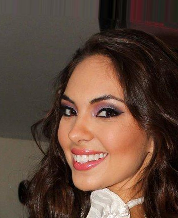} & \includegraphics[width=2.5cm]{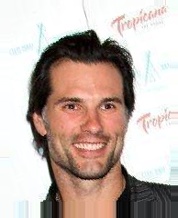} & \includegraphics[width=2.5cm, height=3.05cm]{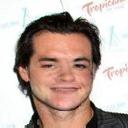} & 
\includegraphics[width=2.5cm]{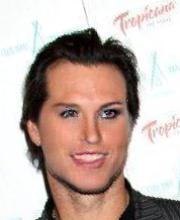}\\
\includegraphics[width=2.5cm]{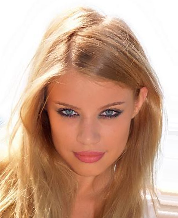} & \includegraphics[width=2.5cm]{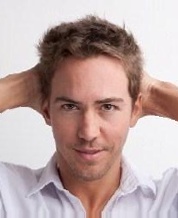} & \includegraphics[width=2.5cm, height=3.05cm]{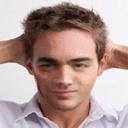} & 
\includegraphics[width=2.5cm]{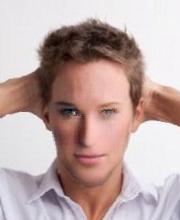}\\
\end{tabular}
\end{adjustbox}
\vspace{-0.3cm}
    \caption{Qualitative results with images from CelebA.}
	\label{fig:qualitative}
\end{figure*}

Figure \ref{fig:qualitative} shows the faces generated by \mname~and the compared method CIAGAN \cite{ciagan_cvpr2020}, with the condition (the 1st column) and source (the 2nd column) images. For the condition identities, we use those that CIAGAN has been trained with for a fair comparison. We can observe that \mname~generates more natural looking images with the facial attributes of the condition face better transferred to the context of the source image. The pose of the source face is better preserved, confirming the effectiveness of the proposed Landmark Attention module that allows for automatic weighting on the full 68 landmarks. 



\vspace{-0.4cm}
\vspace{-1cm}
\section{Conclusions}
In this paper, we proposed \mname, a GAN-based solution for face anonymisation, generating faces with the appearance of any condition image and the facial pose of the source image. Our approach leverages 
landmark-to-landmark geometric reasoning via GCNs to model the relations between the condition and source facial landmarks. We also introduced a landmark attention model to automatically learn the importance of the facial landmarks. We compared our approach with the state-of-the-art approaches both quantitatively and qualitatively, demonstrating a better performance both in term of visual quality and pose preservation. As future work, we will explore alternatives to further improve the face anonymisation performance and adapt the model to operate in real world.

\section*{Acknowledgement}
This work has been supported by the European Union’s Horizon 2020 research and innovation programme under grant agreement No. 957337, and by the European Commission Internal Security Fund for Police under grant agreement No. ISFP-2020-AG-PROTECT-101034216-PROTECTOR.




\newpage
\bibliographystyle{splncs04}
\bibliography{refs}

\end{document}


\maketitle
In the supplementary material, we provide more details on the losses regarding each ablated model presented in the main script. We further provide one more ablation on the weak feature matching loss to justify its importance in training \mname$^{\star}$ (68 lm). 
Finally, we report more qualitative results on both CelebA and LFW, showing a wider range of generated faces with both the advantages and limitations of our proposed method discussed. 

\section{Ablation models and their losses}

In the main script, we presented two ablated models of our proposed approach, i.e. \mname-$LA^{-}$-$D_{id}^{-}$ (29 lm) and \mname-$D_{id}^{-}$ (68 lm). \mname-$LA^{-}$-$D_{id}^{-}$ (29 lm) does not have the landmark attention model and identity discriminator, and its input poses are defined with 29 facial landmarks as in CIAGAN \cite{ciagan_cvpr2020}. \mname-$D_{id}^{-}$ (68 lm) is trained without identity discriminator, but with landmark attention model, thus its input poses have 68 facial landmarks. Moreover, we further provide an ablation model \mname$^{\star}$-$wFM^{-}$ (68 lm) to justify the importance of the weak feature matching (wFM) loss \cite{Chen_2020} for learning the high-level appearance hints. \mname$^{\star}$-$wFM^{-}$ (68 lm) is trained with both landmark attention model and identity discriminator, following a hybrid training strategy indicated by $\star$, however without the wFM loss. 
In the following, we provide more details on the losses employed during the training phase for each of the above-mentioned models.

\textbf{\mname-$LA^{-}$-$D_{id}^{-}$ (29 lm)}
is trained with only image pairs corresponding to the same person. The losses concerning this model include an $L_{1}$ norm with respect to the generated image and the source image, denoted as $L_{Recon}$ in Equation \ref{eq:final_loss_bis}, a perceptual loss $L_{percep}$ between generated and source image on the third layer of a pre-trained VGG network, as well as the gradient penalty loss $L_{GP}$ to prevent exploding gradients, the landmark loss $L_{lm}$ and the classic adversarial loss $L_{GAN}$ that are defined in the main script. 



The complete loss function is defined as:
\begin{equation}
\centering
    L = \lambda_{GAN} L_{GAN} + \lambda_{lm} L_{lm} + \lambda_{Recon} L_{Recon} + \lambda_{percep} L_{percep} + \lambda_{GP} L_{GP},
    \label{eq:final_loss_bis}
\end{equation}
where the weighting parameters are empirically set. Note that the weights differ from Equation 5 in the main script since there is no ID discriminator to be trained and \textit{weak} Feature Matching is not required, given that source and condition images belong to the same person.

\textbf{\mname-$D_{id}^{-}$ (68 lm)} is trained with the same data and losses as \mname-$LA^{-}$-$D_{id}^{-}$ (29 lm), where the only difference lies in the network architecture regarding the use of Landmark Channel Attention. Therefore, the total loss is defined as in Equation \ref{eq:final_loss_bis}.





\textbf{\mname$^{\star}$-$wFM^{-}$ (68 lm)}
is trained with the hybrid dataset, alternating image pairs (the condition image and source image) with the same identity and different identities, to facilitate the appearance learning with the use of the id discriminator. Different from \mname$^{\star}$, we remove the \textit{weak} Feature Matching \cite{Chen_2020} to understand its role on the image generation. 
%
Thus, the total loss function is:
\begin{equation}
\centering
    L = \lambda_{GAN} L_{GAN} + \lambda_{lm} L_{lm} + \lambda_{ID} L_{ID} + \lambda_{Recon} L_{Recon} + \lambda_{GP} L_{GP},
    \label{eq:final_loss}
\end{equation}
where the only difference to the Equation 5 in the main script is the removal of $L_{wFM}$.

\begin{table}[t!] \small
	\centering
	\begin{adjustbox}{max width=0.8\textwidth}
    \begin{tabular}{lccccccc}
    \hline
    \multirow{2}{*}{} & \multirow{2}{*}{$FID \downarrow$} & \multicolumn{2}{c}{Face detection $\uparrow$} & \multicolumn{2}{c}{Face re-identification $\downarrow$} & \multirow{2}{*}{Pose$\downarrow$}\\
    & & dlib & Facenet &  Casia & VGG \\
    \hline
    \mname-$LA^{-}$-$D_{id}^{-}$ (29 lm) & {\bf 5.31} & \textbf{100}\% & \textbf{100}\% & \textbf{2.66}\% & 2.24\%  & \textbf{0.144}\\
    \mname-$D_{id}^{-}$ (68 lm) & 5.79 & \textbf{100}\% & \textbf{100}\% & 4.80\% & 3.06\%  & 0.152\\
    \hline
    \mname$^{\star}$-$wFM^{-}$ (68 lm) $\dagger$ & 9.18 & \textbf{100}\% & \textbf{100}\% & 2.88\% & 2.35\%  & 0.152\\
    \mname$^{\star}$ (68 lm) & 6.49 & \textbf{100}\% & \textbf{100}\% & 2.68\% & \textbf{1.88}\%  & 0.150\\
    \hline
\end{tabular}
\end{adjustbox}
\caption{Performance of the proposed \mname~and its variants evaluated on the test set of CelebA. The model denoted with $\dagger$ is not included in the main ablation study.}
\label{tab:comparison_celeba}
\end{table}

As reported in Table \ref{tab:comparison_celeba}, we can observe that \textit{weak} Feature Matching loss is important for the network to generate more natural-looking faces as indicated by the FID score. It plays a fundamental role during the training epochs when the images of the condition and source are of different identities. The face de-identification performance of \mname$^{\star}$-$wFM^{-}$ (68 lm) is also worse than that of \mname$^{\star}$ (68 lm).







\section{Qualitative Results}
\begin{figure*}[!ht]
   \centering
   \begin{adjustbox}{max width=1.\textwidth}
\begin{tabular}{c|c|c|c|c|c|c}
Condition&Source&CIAGAN&\mname-$LA^{-}$-$D_{id}^{-}$ & \mname-$D_{id}^{-}$ & \mname$^{\star}$-$wFM^{-}$ & \mname$^{\star}$ \\
&
\includegraphics[width=2.5cm]{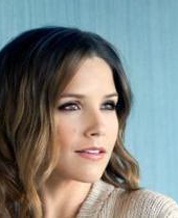}&
\includegraphics[width=2.5cm, height=3.05cm]{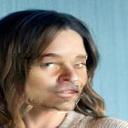}&
\includegraphics[width=2.5cm]{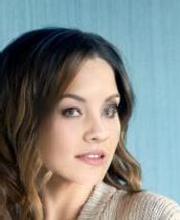}&
\includegraphics[width=2.5cm]{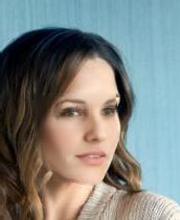}&
\includegraphics[width=2.5cm]{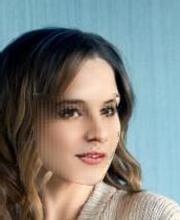}&
\includegraphics[width=2.5cm]{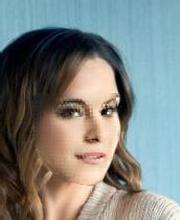}\\
&\includegraphics[width=2.5cm]{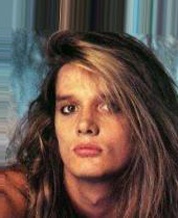}&
\includegraphics[width=2.5cm, height=3.05cm]{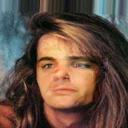}&
\includegraphics[width=2.5cm]{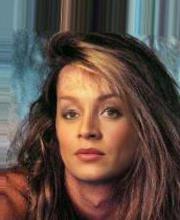}&
\includegraphics[width=2.5cm]{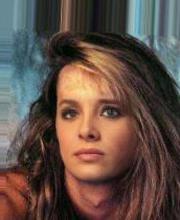}&
\includegraphics[width=2.5cm]{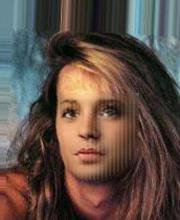}&
\includegraphics[width=2.5cm]{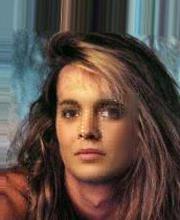}\\
\includegraphics[width=2.5cm]{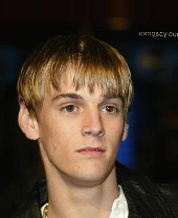}&
\includegraphics[width=2.5cm]{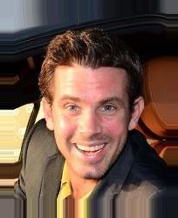}&
\includegraphics[width=2.5cm, height=3.05cm]{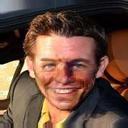}&
\includegraphics[width=2.5cm]{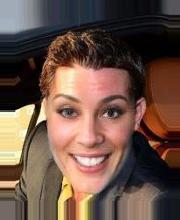}&
\includegraphics[width=2.5cm]{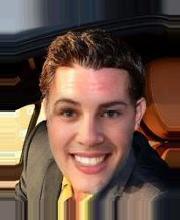}&
\includegraphics[width=2.5cm]{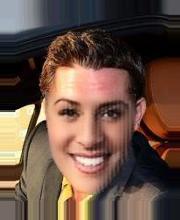}&
\includegraphics[width=2.5cm]{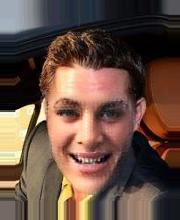}\\
&\includegraphics[width=2.5cm]{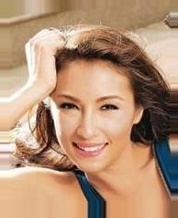}&
\includegraphics[width=2.5cm, height=3.05cm]{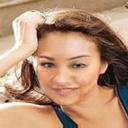}&
\includegraphics[width=2.5cm]{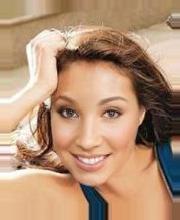}&
\includegraphics[width=2.5cm]{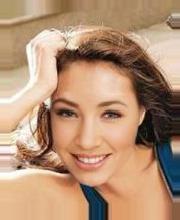}&
\includegraphics[width=2.5cm]{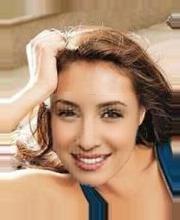}&
\includegraphics[width=2.5cm]{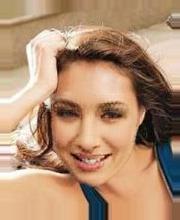}\\
&\includegraphics[width=2.5cm]{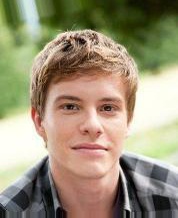}&
\includegraphics[width=2.5cm, height=3.05cm]{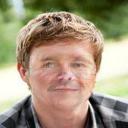}&
\includegraphics[width=2.5cm]{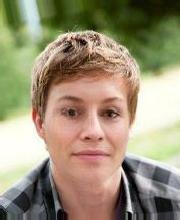}&
\includegraphics[width=2.5cm]{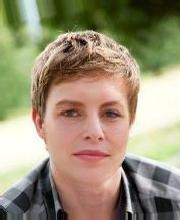}&
\includegraphics[width=2.5cm]{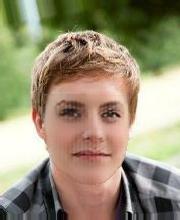}&
\includegraphics[width=2.5cm]{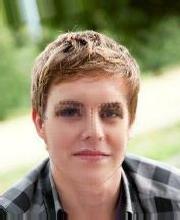}\\

\end{tabular}
\end{adjustbox}
    \caption{More qualitative results on CelebA}
    \label{fig:supp_qualitative_2} 
\end{figure*}

\begin{figure*}[!ht]
   \centering
   \begin{adjustbox}{max width=1.\textwidth}
\begin{tabular}{c|c|c|c|c|c|c}
Condition&Source&CIAGAN&\mname-$LA^{-}$-$D_{id}^{-}$ & \mname-$D_{id}^{-}$ & \mname$^{\star}$-$wFM^{-}$ & \mname$^{\star}$ \\
&
\includegraphics[width=2.5cm]{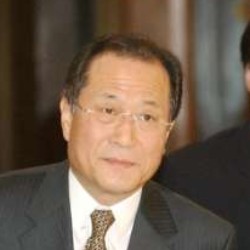}&
\includegraphics[width=2.5cm]{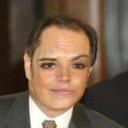}&
\includegraphics[width=2.5cm]{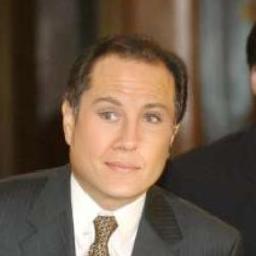}&
\includegraphics[width=2.5cm]{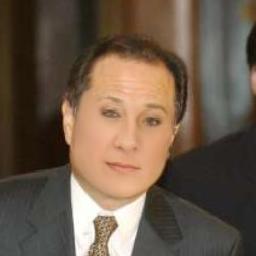}&
\includegraphics[width=2.5cm]{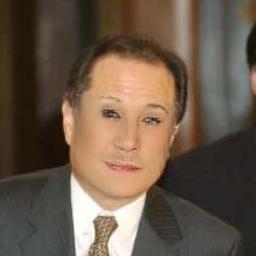}&
\includegraphics[width=2.5cm]{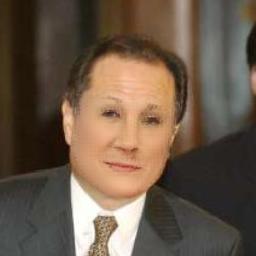}\\
&\includegraphics[width=2.5cm]{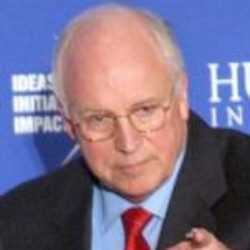}&
\includegraphics[width=2.5cm]{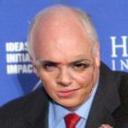}&
\includegraphics[width=2.5cm]{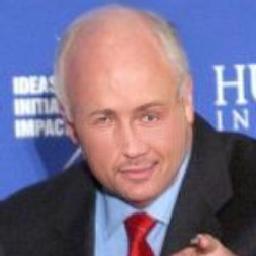}&
\includegraphics[width=2.5cm]{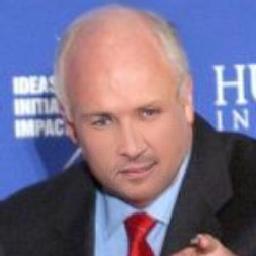}&
\includegraphics[width=2.5cm]{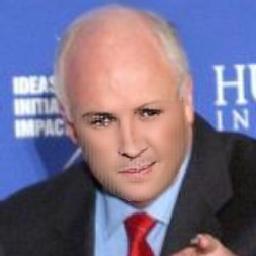}&
\includegraphics[width=2.5cm]{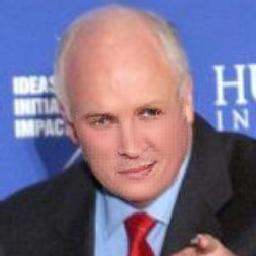}\\
\includegraphics[width=2.05cm]{Figures/supp-celeba/032486}&
\includegraphics[width=2.5cm]{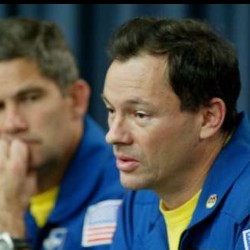}&
\includegraphics[width=2.5cm]{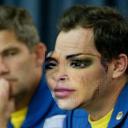}&
\includegraphics[width=2.5cm]{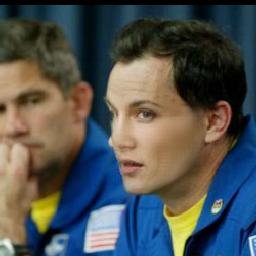}&
\includegraphics[width=2.5cm]{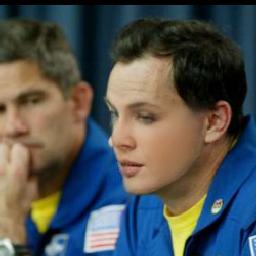}&
\includegraphics[width=2.5cm]{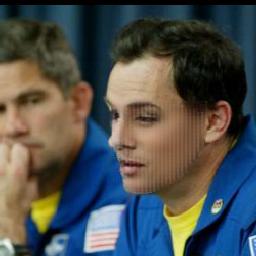}&
\includegraphics[width=2.5cm]{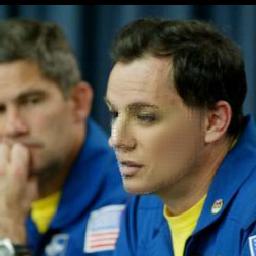}\\
&\includegraphics[width=2.5cm]{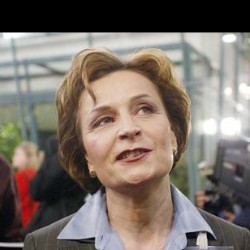}&
\includegraphics[width=2.5cm]{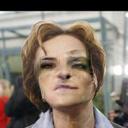}&
\includegraphics[width=2.5cm]{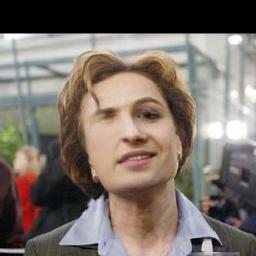}&
\includegraphics[width=2.5cm]{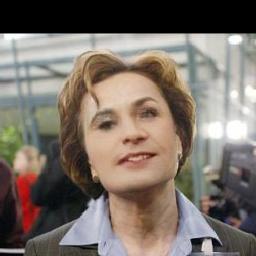}&
\includegraphics[width=2.5cm]{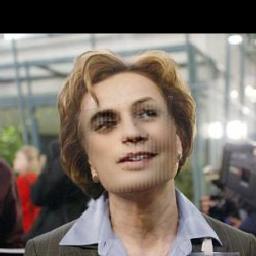}&
\includegraphics[width=2.5cm]{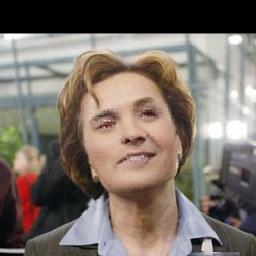}\\
&\includegraphics[width=2.5cm]{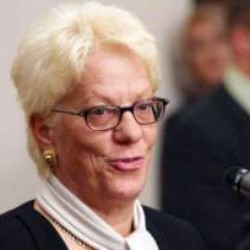}&
\includegraphics[width=2.5cm]{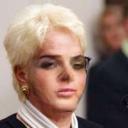}&
\includegraphics[width=2.5cm]{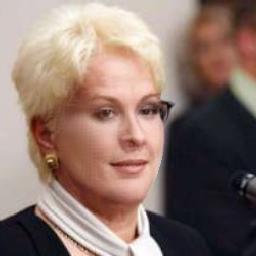}&
\includegraphics[width=2.5cm]{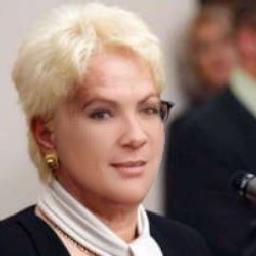}&
\includegraphics[width=2.5cm]{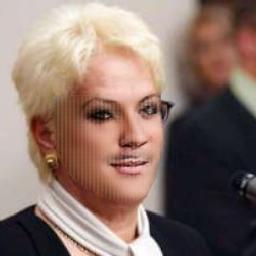}&
\includegraphics[width=2.5cm]{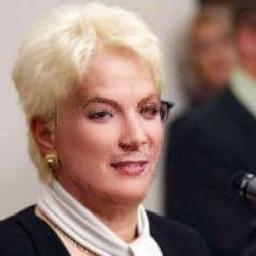}\\
\end{tabular}
\end{adjustbox}
    \caption{Qualitative results on LFW}
    \label{fig:supp_qualitative_lfw_1} 
\end{figure*}

We report more qualitative results produced with faces in CelebA (as shown in Fig. \ref{fig:supp_qualitative_2}) and LFW (as shown in Fig. \ref{fig:supp_qualitative_lfw_1}). 
%
Our proposed method is able to generate more natural faces with better facial pose preserved, although in some circumstances the generated faces appear similar to the source images. CIAGAN, on the other hand, is able to generate better anonymised face, however the generated faces are of lower visual quality. Moreover, from Fig. \ref{fig:supp_qualitative_lfw_1} (the last row), we notice that methods including CIAGAN and ours, which generate faces only on the masked facial area for anonymising the faces, present artefacts with facial accessories, e.g. glasses, that are out of the face area.

\clearpage

\bibliography{egbib}